%% file: main.tex
\documentclass[sigconf]{acmart}
\AtBeginDocument{%
  }

\copyrightyear{2026}
\acmYear{2026}
\setcopyright{cc}
\setcctype{by}
\acmConference[ICMR '26]{International Conference on Multimedia Retrieval}{June 16--19, 2026}{Amsterdam, Netherlands}
\acmBooktitle{International Conference on Multimedia Retrieval (ICMR '26), June 16--19, 2026, Amsterdam, Netherlands}
\acmDOI{10.1145/3805622.3810887}
\acmISBN{979-8-4007-2617-0/2026/06}

\usepackage{multirow}  
\usepackage{array}
\usepackage{amsfonts}
\usepackage{stackengine} 
\usepackage{makecell} 
\usepackage{longtable}
\usepackage{tabularx}
\usepackage{fontawesome}
\usepackage{cuted}  
\usepackage{placeins}
\newcolumntype{C}{>{\centering\arraybackslash}X}

\usepackage{subcaption}           
\usepackage{booktabs}             
\usepackage{colortbl}             

\usepackage{amssymb}              

\hypersetup{
  colorlinks=true,
  citecolor=blue,
  linkcolor=blue,
  urlcolor=blue
}
\usepackage{bbm}

\begin{document}

\title{SparseStreet: Sparse Gaussian Splatting for Real-Time \\
Street Scene Simulation}



\author{Qingpo Wuwu}
\affiliation{%
  \institution{Peking University}
  \country{China}
}
\email{2401112105@stu.pku.edu.cn}

\author{Xiaobao Wei}
\affiliation{%
  \institution{Chinese Academy of Sciences}
  \country{China}
}
\email{weixiaobao0210@gmail.com}

\author{Peng Chen}
\affiliation{%
  \institution{Chinese Academy of Sciences}
  \country{China}
}
\email{chenpeng23@mails.ucas.ac.cn}

\author{Nan Huang}
\affiliation{%
  \institution{University of Illinois Urbana-Champaign}
  \country{United States}
}
\email{nanh3@illinois.edu}

\author{Zhongyu Zhao}
\affiliation{%
  \institution{Peking University}
  \country{China}
}
\email{zhaozhongyu2000@pku.edu.cn}

\author{Hao Wang}
\affiliation{%
  \institution{Peking University}
  \country{China}
}
\email{haowang@stu.pku.edu.cn}

\author{Ming Lu}
\affiliation{%
  \institution{Peking University}
  \country{China}
}
\email{lu199192@gmail.com}

\author{Ningning Ma}
\affiliation{%
  \institution{Autonomous Driving Development, NIO}
  \country{China}
}
\email{mnn.thu@gmail.com}

\author{Shanghang Zhang}
\affiliation{%
  \institution{Peking University}
  \country{China}
}
\email{shanghang@pku.edu.cn}
\renewcommand{\shortauthors}{Wuwu et al.}



\input{sec/1_main/00_abstract}

\ccsdesc[500]{Computing methodologies~Reconstruction}
\ccsdesc[300]{Computing methodologies~Rendering}
\ccsdesc[300]{Computing methodologies~Supervised learning}

\keywords{Gaussian Splatting, Real-time rendering, Computer graphics}

\begin{teaserfigure}
  \centering
  \includegraphics[width=\textwidth]{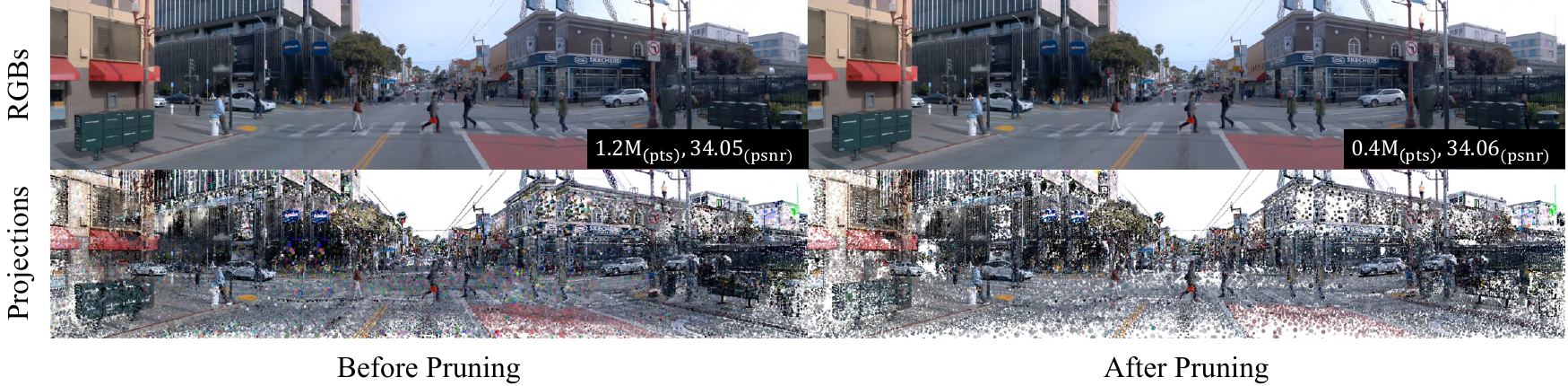}
  \caption{Visualization of Gaussian projections before and after pruning across three camera views.
  The first row displays the RGB images, while the second row shows the projected Gaussian points.}
  \label{fig:teaser}
\end{teaserfigure}

\maketitle
\input{sec/1_main/01_intro}

\input{sec/1_main/02_related_work}

\input{sec/1_main/03_preliminaries}

\input{sec/1_main/04_method}

\input{sec/1_main/05_experiment}

\input{sec/1_main/06_conclusion}

\input{sec/1_main/10_acknowledgments}
\input{sec/1_main/11_reference}

\end{document}

%% file: sec/1_main/00_abstract.tex
\begin{abstract}
While 3D Gaussian Splatting has shown promising results in street scene reconstruction, existing methods require massive numbers of Gaussian primitives to capture fine details, leading to prohibitive storage costs and slow rendering speeds. We observe that dynamic objects (e.g., vehicles and pedestrians) demand high-fidelity representations to maintain temporal consistency, while static background regions often contain substantial redundancy. Motivated by this, we propose \textbf{SparseStreet}, a general compression framework specifically designed for street scenes. First, we introduce a node-based learnable pruning strategy that systematically removes low-contributing Gaussian primitives while preserving visually critical regions. Second, after the scene representation stabilizes, we apply background compression, further reducing redundancy in static regions. Our method effectively preserves the geometry and appearance of dynamic objects while significantly reducing the total number of Gaussian primitives. Extensive experiments on the Waymo and nuScenes demonstrate that SparseStreet achieves up to 80\% compression ratio with minimal quality degradation, enabling resource-efficient, high-fidelity dynamic scene reconstruction. Project website: \url{https://sparsestreet.github.io/}.

\end{abstract}

%% file: sec/1_main/01_intro.tex

\section{Introduction}
\label{sec:intro}
Photorealistic reconstruction of dynamic scenes is fundamental to autonomous driving applications, including simulation~\cite{wang2025plgs, wang2025roboarmgs}, testing~\cite{cao2026evodrivevla, cao2026fastdrivevla}, and perception system validation~\cite{zeng2025rethinking, wang2025embodiedocc}. While traditional simulators such as NVIDIA DRIVE Sim~\cite{nvidia_drive_sim} and CarSim~\cite{carsim} offer controlled environments, they often lack realism and require extensive manual effort, as artist-generated assets have limited scale and diversity. The emergence of neural rendering techniques, particularly 3D Gaussian Splatting (3DGS)~\cite{2023_8_08-3dgs_for_real_time_radiance_field_rendering, wei2025gazegaussian, wei2025graphavatar, chen2025mixedgaussianavatar, chen2026rsatalker} and Neural Radiance Fields (NeRF)~\cite{2020_08_03-NeRF-Representing_Scenes_as_Neural_Radiance_Fields_for_View_Synthesis, wei2024nto3d}, has transformed scene reconstruction with their ability to capture complex geometries and appearances. Extensions of these methods to dynamic scenes incorporate time dimensions and deformation networks to model rigid and non-rigid motions~\cite{2020_11_25-Nerfies-Deformable_Neural_Radiance_Fields, 2020_11_27-D-NeRF-Neural_Radiance_Fields_for_Dynamic_Scenes, 2020_11_26-Neural_Scene_Flow_Fields_for_Space-Time-View_Synthesis_of_Dynamic_Scenes, 2020_12_22-Non-Rigid_Neural_Radiance_Fields-Reconstruction_and_Novel_View_Synthesis_of_a_Dynamic_Scene_From_Monocular_Video, 2021_6_24-HyperNeRF-A_Higher_Dimensional_Representation_for_Topologically_Varying_Neural_Radiance_Fields, 2022_11_25-Space-Time_Neural_Irradiance_Fields_for_Free-Viewpoint_Video, 2021_10_26-H-NeRF-Neural_Radiance_Fields_for_Rendering_and_Temporal_Reconstruction_of_Humans_in_Motion, 2022_2_17-Fourier_PlenOctrees_for_Dynamic_Radiance_Field_Rendering_in_Real-time, 2022_10_17-DANOs-Differentiable_Physics_Simulation_of_Dynamics-Augmented_Neural_Objects, li2026manipdreamer3d, wei2025emd, wei2026parkgaussian}. However, applying these techniques to autonomous driving scenarios remains challenges due to the complexity and scale of urban environments, which involve both static backgrounds and highly dynamic objects such as vehicles and pedestrians.

To address these challenges, researchers have developed specialized adaptations of 4D-GS and dynamic NeRF frameworks to handle both static and dynamic elements in driving scenes. These methods generally fall into two categories: supervised approaches and self-supervised approaches. Supervised methods rely on auxiliary signals, such as segmentation masks from SAM~\cite{xie2021_segformer, 2023_4_SAM_segment_anything, wei2024medsam}, depth estimations from DepthAnything~\cite{2024_01_15_depthanything}, combined depth and optical flow from Dynamo~\cite{2023_10_29-dynamo_depth}, or 3D bounding boxes from annotated datasets. Building upon these supervised signals, \cite{2020_11_20-neural_scene_graph} introduces the concept of scene graphs which decompose scenes into hierarchical structures with dynamic actors and static backgrounds as distinct nodes, connected with edges encoding transformation parameters that represent motion over time. Recent works in this category include StreetGaussian \cite{2024_01_02-street_gaussian-modelling_dynamic_urban_scenes_with_gs, 2024_3_19-Hugs}, which employs bounding box annotations for precise object localization and motion modeling, and OmniRe~\cite{2024_8_29-omnire}, which further advances this paradigm through detailed representation of human actors and non-rigid objects. Conversely, self-supervised methods \cite{2024_3_19-HUGS-Holistic_Urban_3D_Scene_Understanding_via_Gaussian_Splatting, 2024_6_26-VDG-Vision-Only_Dynamic_Gaussian_for_Driving_Simulation, 2022_3_31-D2NeRF-Self_Supervised_Decoupling_of_Dynamic_and_Static_Objects_from_a_Monocular_Video, 2023_3_25-SUDS-Scalable_Urban_Dynamic_Scenes, 2023_11_3-EMERNERF-EMERGENT_SPATIAL_TEMPORAL_SCENE_DECOMPOSITION_VIA_SELF-SUPERVISION, 2024_3_20-PVG-periodic_vibration_gaussian-dynamic_urban_scene_reconstruction, 2024_5_30-S3Gaussian-Self_Supervised_Street_Gaussians_for_Autonomous_Driving,peng2024_desire_gs} eliminate the need for annotations by exploiting temporal dynamics within scenes.


Despite their impressive visual quality, existing methods face significant limitations in storage and rendering efficiency. \textbf{3DGS}-based approaches often require millions of Gaussian primitives to reconstruct a single scene, resulting in substantial computational and storage burdens that hinder their deployment in real-world applications. To address the compression challenges in dynamic street scenes, we propose \textbf{SparseStreet}, a general compression framework that can be easily integrated into existing scene graph-based reconstruction methods (Fig.~\ref{fig:method_overview}). Our approach improves compression efficiency by incorporating node-aware pruning and background compression. Specifically, we enhance each Gaussian primitive with learnable masking scores to capture its contribution to the scene and design a node-aware regularization strategy that applies different pruning strengths based on scene graph nodes. Additionally, we further compress background regions by computing global importance metrics that combine blending weights with projected areas. This design allows for more efficient representation of complex street scenes with varying compression requirements.


We validate our approach through extensive experiments on the Waymo \cite{sun2020_waymo} and nuScenes \cite{caesar2020nuscenes} datasets, by integrating \textbf{SparseStreet} with representative methods: StreetGS and OmniRe. Our main contributions include:

\begin{itemize}
\item We propose SparseStreet, a plug-and-play compression framework that efficiently reduces redundancy in existing street scene reconstruction methods.

\item We demonstrate the broad applicability of SparseStreet by successfully integrating it with state-of-the-art neural scene graph methods, showing consistent improvements across different reconstruction approaches.

\item Comprehensive evaluations on the Waymo and nuScenes datasets demonstrate that our approach achieves up to 80\% reduction in Gaussian primitives while maintaining visual fidelity of dynamic objects and improving rendering speed by 2× FPS.
\end{itemize}

%% file: sec/1_main/02_related_work.tex
\section{Related Work}

\noindent\textbf{Autonomous Driving Simulation. }
High-fidelity simulation of street scenes is a cornerstone for developing and validating autonomous driving systems. Traditional simulators such as CARLA~\cite{2017_11_10-CARLA-An_Open_Urban_Driving_Simulator} and AirSim~\cite{2017_7_18-AirSim-High_Fidelity_Visual_and_Physical_Simulation_for_Autonomous_Vehicles} rely heavily on handcrafted assets and rule-based environments, which limit realism and scalability. To address these limitations, neural scene representation has emerged as a promising alternative. Neural Radiance Fields (NeRF)~\cite{2020_08_03-NeRF-Representing_Scenes_as_Neural_Radiance_Fields_for_View_Synthesis} pioneered this direction, with follow-up works~\cite{2020_11_20-NSG-Neural_Scene_Graphs_for_Dynamic_Scenes, 2021_12_20-Mega-NERF_Scalable_Construction_of_Large-Scale_NeRFs_for_Virtual_Fly_Through, 2021_11_29-Urban_Radiance_Fields, 2023_5_2-NFL-Neural_LiDAR_Fields_for_Novel_View_Synthesis} extending its capabilities to large-scale urban scenes. Building upon these foundations, subsequent research~\cite{2022_2_10-Block-NeRF-Scalable_Large_Scene_Neural_View_Synthesis, 2023_3_25-SUDS-Scalable_Urban_Dynamic_Scenes, 2023_7_20-Urban_Radiance_Field_Representation_with_Deformable_Neural_Mesh_Primitives, 2023_11_9-RealTime_Neural_Rasterization_for_Large_Scenes} improve rendering efficiency and scalability. To better model dynamic environments, several methods incorporate scene graphs to decompose static backgrounds and dynamic agents~\cite{2020_11_20-NSG-Neural_Scene_Graphs_for_Dynamic_Scenes, 2023_7_27-MARS-An_Instance_aware_Modular_and_Realistic_Simulator_for_Autonomous_Driving, 2023_11_26-NeuRAD-Neural_Rendering_for_Autonomous_Driving, 2024_3_29-Multi-Level_Neural_Scene_Graphs_for_Dynamic_Urban_Environments}. With the advent of 3D Gaussian Splatting (3DGS)~\cite{2023_8_08-3dgs_for_real_time_radiance_field_rendering}, NeRF-based techniques have also been adapted to the 3DGS framework. DrivingGaussian~\cite{2024_2_27-drivinggaussian-compusing_gaussian_splatting_for_surronding_dynamic_autonomous_driving_scenes}, StreetGaussian~\cite{2024_01_02-street_gaussian-modelling_dynamic_urban_scenes_with_gs}, and OmniRe~\cite{2024_8_29-omnire} adopt scene graph structures for supervised dynamic scene reconstruction. To eliminate the need for costly annotations, self-supervised approaches such as S3Gaussian~\cite{2024_5_30-S3Gaussian-Self_Supervised_Street_Gaussians_for_Autonomous_Driving}, DeSiRe-GS~\cite{peng2024_desire_gs}, and PVG~\cite{2024_3_20-PVG-periodic_vibration_gaussian-dynamic_urban_scene_reconstruction} leverage temporal cues for unsupervised static-dynamic separation.

Despite their effectiveness, previous street Gaussian splatting methods overlook the significant storage demands, limiting their practical applicability. To address this issue, we introduce SparseStreet, a compression framework specifically designed to reduce redundancy in street Gaussian splatting. 

\noindent\textbf{Compression Techniques for 3D Gaussian Splatting. }
As 3D-GS has revealed critical challenges in storage efficiency, with scenes often requiring hundreds of millions of Gaussians to capture fine details, researchers have developed various compression approaches to address these scalability issues. One prominent line of work focuses on reducing the number of Gaussian primitives while preserving scene fidelity. For instance, LightGaussian~\cite{fan_2023_11_28-lightgaussian} introduced a dual-focus compression strategy that reduces both Gaussian count and feature dimensionality, achieving substantial compression with minimal quality loss. Building on this foundation, C3DGS~\cite{niedermayr_2024_1_22-c3dgs} advanced the field with sensitivity-aware clustering and fine-tuning, demonstrating significant storage reduction through targeted parameter compression. In parallel, complementary methods have tackled other aspects of Gaussian representation. EAGLES~\cite{girish_2024_9_26_eagles} prioritized compression of memory-intensive attributes, such as color and rotation, while Compact-3DGS~\cite{lee2024_5_2_compact} exploited scene redundancy by organizing parameters into locally homogeneous 2D grids combined with learnable masking and residual vector quantization. Other approaches have reconceptualized the problem entirely: Mini-Splatting~\cite{fang_2024_3_21-mini_splatting} reorganized the spatial distribution of Gaussians to enhance storage efficiency, and Ye et al.~\cite{ye2024_Fragment_pruning} innovated at the fragment level, selectively pruning fragments to accelerate rendering.

While these methods have demonstrated significant success in compressing general static scenes, they have not been extended to dynamic driving scenes, which present unique challenges due to the coexistence of redundant static backgrounds and dynamic objects requiring higher fidelity representation.

%% file: sec/1_main/03_preliminaries.tex
\input{sec/2_figures/04_method/1_method_overview}

\section{Preliminaries}
\label{sec:Preliminaries} 
\subsection{3D Gaussian Splatting}
\label{subsec:3dgs}

The 3D-GS framework \cite{2023_8_08-3dgs_for_real_time_radiance_field_rendering} represents scenes using a set of Gaussian primitives denoted as $\mathbb{G}=\left\{\left(\mu_k, \Sigma_k, \alpha_k, \mathbf{c}_k\right)\right\}_{k=1}^K$, where $K$ indicates the total primitive count. Each Gaussian is characterized by its center position $\mu_k \in \mathbb{R}^3$, covariance matrix $\Sigma_k \in \mathbb{R}^{3 \times 3}$ defining its spatial extent and orientation, opacity value $\alpha_k \in [0,1]$, and color information $\mathbf{c}_k$. The density distribution of each Gaussian follows:
\begin{equation}
G_k(x)=e^{-\frac{1}{2}(x-\mu_k)^T \Sigma_k^{-1}(x-\mu_k)},
\end{equation}
where $x$ represents any 3D world coordinate.

\noindent\textbf{Rendering Methodology.}
To generate an image, the rendering pipeline projects each 3D Gaussian onto the image plane. This projection transforms the 3D center $\mu_k$ to a 2D position $\mu_k^{2D}$ and converts the world-space covariance $\Sigma_k$ to screen space via $\Sigma_k^{\prime}=JW\Sigma_k W^TJ^T$, where $W$ represents the viewing transformation and $J$ denotes the projective transformation Jacobian. The color at any screen position $x$ is computed through alpha blending:
\begin{equation}
C(x)=\sum_{k \in \mathcal{N}(\mathbf{x})} \mathbf{c}_k \alpha_k(x) \prod_{j=1}^{k-1}\left(1-\alpha_j(x)\right),
\end{equation}
with $\alpha_k(x)=\alpha_k \exp \left(-\frac{1}{2}\left(x-\mu_k^{2D}\right)^T {\Sigma_k^{\prime}}^{-1}\left(x-\mu_k^{2D}\right)\right)$ representing each Gaussian's contribution to pixel $x$, and $\mathcal{N}(\mathbf{x})$ indicating the set of Gaussians affecting this pixel.

\subsection{Scene Graph Representation}
\label{subsec:scene_graph}

A Scene Graph \cite{2020_11_20-neural_scene_graph} is a hierarchical representation that organizes scene components into nodes, enabling structured modeling of static and dynamic elements. It comprises a Background Node for static elements (roads, buildings), Rigid Nodes for vehicles, Deformable Nodes for non-rigid objects, and SMPL Nodes \cite{loper2023_smpl} for humans. Each node type is associated with specific Gaussian primitives. For dynamic scenes, Rigid nodes maintain consistent Gaussian attributes while their positions evolve through $SE(3)$ transformations. Deformable and SMPL nodes handle local deformations using networks that predict temporal changes in Gaussian attributes.

%% file: sec/2_figures/04_method/1_method_overview.tex
\begin{figure*}[t!]
\centering
\includegraphics[width=\textwidth]{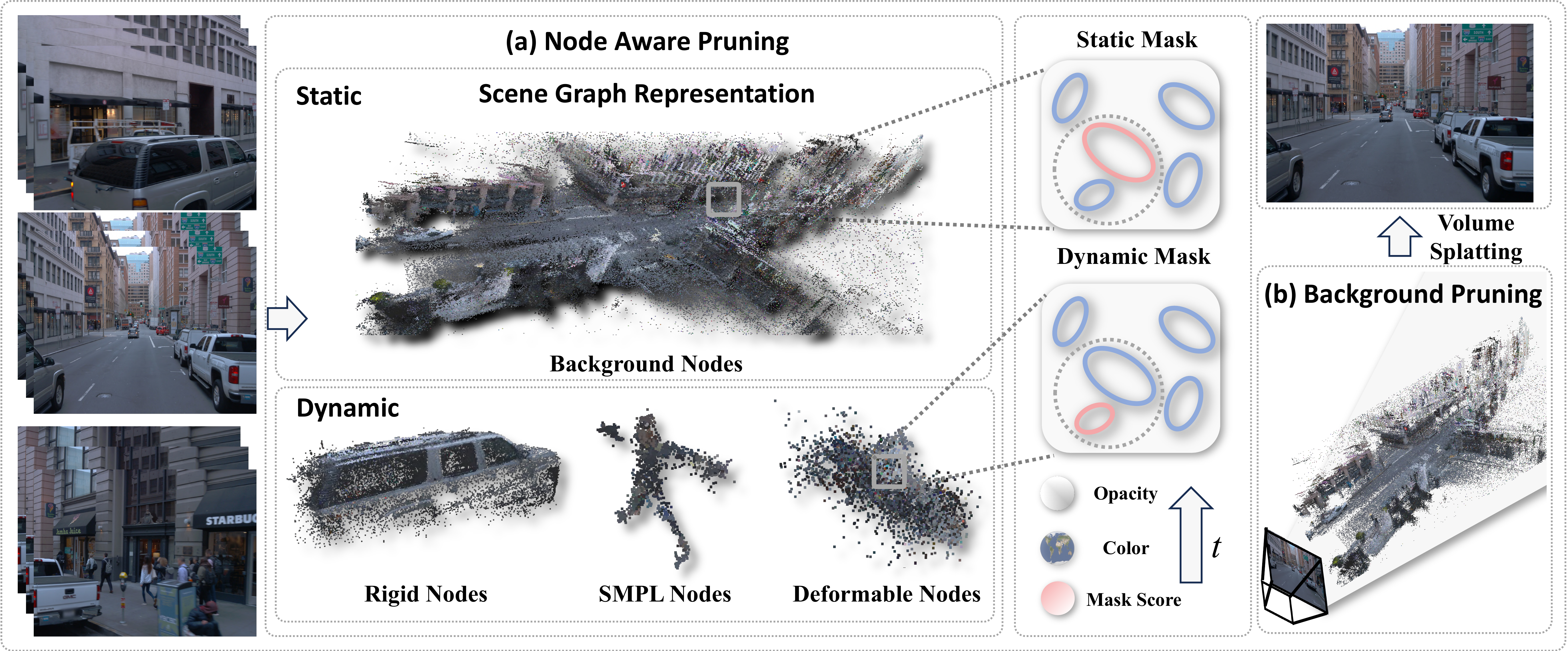}
\captionof{figure}{Overview of SparseStreet. Given the street video as input, our method first constructs a hierarchical scene graph representation. Then we introduce a two-stage compression technique to prune redundant Gaussian primitives: (a) Node-aware Pruning, which applies dynamic masks for temporally varying objects and static masks for persistent elements based on different node types; and (b) Background Compression, which focuses on pruning redundant Gaussians in static background based on their global importance.}
\label{fig:method_overview}
\end{figure*}

%% file: sec/1_main/04_method.tex
\section{Methodology}

We propose SparseStreet, as illustrated in Fig.~\ref{fig:method_overview}, a compression framework tailored for dynamic autonomous driving scenes.

\subsection{Problem Formulation}
Given a set of Gaussian primitives $\mathbb{G} = \left\{\left(\mu_k, \Sigma_k, \alpha_k, \mathbf{c}_k\right)\right\}_{k=1}^K$, where $K$ represents the total number of primitives, our goal is to reduce the number of primitives while maintaining high rendering quality and preserving perceptual fidelity. Specifically, we aim to learn a binary mask $M_k \in \{0, 1\}$ for each Gaussian primitive, which determines whether $G_k$ is preserved ($M_k=1$) or removed ($M_k=0$). The pruning process can be formulated as:

\begin{equation}
\hat{\mathbb{G}} = \left\{ G_k \mid M_k = 1, \forall k \in \{1, \dots, K\} \right\},
\end{equation}
where $\hat{\mathbb{G}}$ denotes the pruned set of Gaussian primitives.

\input{sec/3_tables/05_experiment/1_comparison_waymo}

\subsection{Node-aware Pruning}
To reduce the number of Gaussian primitives while maintaining reconstruction quality, we enhance each Gaussian primitive with an additional learnable mask attribute $m_k \in \mathbb{R}$, where $k$ indexes the $K$ total primitives in the scene. Based on $m_k$, a binary mask $M_k \in \{0, 1\}$ is generated using the straight-through estimator~\cite{bengio2013_STE_straight_through_estimator}, which allows gradient flow during optimization. This binary mask is formulated as:

\begin{equation}
M_k = \text{sg}(\mathbbm{1}[\sigma(m_k) > \epsilon] - \sigma(m_k)) + \sigma(m_k),
\end{equation}
where $\epsilon$ is the masking threshold, $\text{sg}(\cdot)$ is the stop-gradient operator, $\mathbbm{1}[\cdot]$ is the indicator function, and $\sigma(\cdot)$ is the sigmoid function. 

The binary mask $M_k$ is applied to both the scale and opacity attributes of each Gaussian to determine its contribution to the scene. The covariance matrix $\Sigma_k$ and the opacity $\alpha_k(x)$ are reformulated as:

\begin{equation}
\hat{\Sigma}_k = R(\mathbf{q}_k)S(M_k \mathbf{s}_k)S(M_k \mathbf{s}_k)^T R(\mathbf{q}_k)^T,
\end{equation}

\begin{equation}
\hat{\alpha}_k(x) = M_k \alpha_k \exp{\left(-\frac{1}{2}(x - \mu_k^{2D})^T \hat{\Sigma}_k^{\prime -1}(x - \mu_k^{2D})\right)},
\end{equation}
where $\hat{\Sigma}_k^{\prime}$ denotes the 2D projected covariance matrix after masking, and $M_k$ directly modulates the Gaussian’s contribution to both its spatial extent and transparency.

\noindent\textbf{Node-aware Regularization. }  
In dynamic scenes, applying a uniform pruning strength across all nodes often leads to severe degradation, especially for nodes representing dynamic objects such as vehicles or pedestrians. To address this, we adjust pruning strength based on the scene graph node associated with each Gaussian by assigning different regularization coefficients $\lambda_n$ to each node $n$. The masking loss $\mathcal{L}_{\text{mask}}$ is defined as:

\begin{equation}
\mathcal{L}_{\text{mask}} = \frac{1}{K} \sum_{n \in \mathcal{N}} \lambda_n \sum_{k \in \mathcal{G}_n} \sigma(m_k),
\label{eq:mask_regularization}
\end{equation}
where $\mathcal{N}$ denotes the set of all nodes in the scene graph, $\mathcal{G}_n$ represents the set of Gaussians associated with node $n$, and $\sigma(m_k)$ represents the likelihood of retaining Gaussian $G_k$ during the pruning process. Dynamic nodes are assigned smaller $\lambda_n$ values to preserve their Gaussians, while static nodes receive larger $\lambda_n$ values for aggressive pruning.

\noindent\textbf{Time-Dependent Mask Modeling.} 
In dynamic scenes, certain Gaussians only contribute to rendering during specific time intervals. Yet static mask scoring treats all Gaussians uniformly across time. To address this, we introduce time-dependent mask scores that adapt to temporal visibility patterns.

For Gaussians that exhibit temporal variations, we replace the static mask parameter $m_k$ with a time-dependent function:
\begin{equation}
m_k(t, \mathbf{p}_k) = \text{MLP}(t, \mathbf{f}_k^t, \mathbf{p}_k),
\end{equation}

where $t \in [0,1]$ represents normalized time, $\mathbf{f}_k^t \in \mathbb{R}^{d}$ denotes learnable time-specific features for Gaussian $k$, $\mathbf{p}_k$ is the normalized 3D position, and MLP is a multilayer perceptron followed by a sigmoid activation.

The time-dependent masking loss is reformulated as:

\begin{equation}
\mathcal{L}_{\text{mask}}^t = \frac{1}{K} \sum_{n \in \mathcal{N}} \lambda_n \sum_{k \in \mathcal{G}_n} \frac{1}{T} \sum_{t=1}^T \sigma(m_k(t, \mathbf{p}_k)),
\end{equation}

where $T$ represents the number of sampled time steps during training. This formulation ensures that Gaussians are only penalized during their active time periods, preventing over-pruning of temporally sparse dynamic objects while maintaining compression effectiveness for persistent elements.

\noindent\textbf{Training. }  
We train the Gaussian parameters to represent the scene for each time frame. Following scene graph-based reconstruction methods, the overall optimization objective is defined as:

\begin{equation}
\mathcal{L}_{\text{total}} = \mathcal{L}_{\text{recon}} + \lambda_{\text{m}} \mathcal{L}_{\text{mask}},
\end{equation}
where $\mathcal{L}_{\text{recon}}$ is the reconstruction loss tailored to the scene graph-based reconstruction method, and $\mathcal{L}_{\text{mask}}$ is the masking loss defined in Eq. \ref{eq:mask_regularization}. This formulation ensures that our method can be seamlessly integrated as a plug-and-play module.

\input{sec/3_tables/05_experiment/2_comparison_nuscenes}

\subsection{Background Pruning}

Static background regions often contain significant redundancy and can be effectively reconstructed using a relatively small number of Gaussian primitives, as shown in Fig.~\ref{fig:2_ground_comparisons}. Based on this observation, we further compress static background regions by removing redundant Gaussian primitives using importance-based metrics.

We adopt importance-based pruning that utilizes blending weights to evaluate the contribution of each Gaussian. For each rendered image $m$, the importance is computed by combining blending weights with the projected area of Gaussians:

\begin{equation}
I_i^{(m)} = \sum_{j=1}^{K_m} \frac{w_{ij}^{(m)}}{S_i^{(m)}}, \quad I_i = \sum_{m=1}^M I_i^{(m)} \cdot \delta(i \in I_{\text{max}}^{(m)}),
\end{equation}

where $S_i^{(m)}$ is the 2D projected area of $G_i$ on image $m$, $K_m$ is the total number of rays intersecting with $G_i$ in image $m$, $\delta(\cdot)$ is an indicator function, and $I_{\text{max}}^{(m)}$ represents the set of indices for Gaussians contributing the most to the rendered image $m$.

\noindent\textbf{Challenges in Dynamic Scene Pruning.} Applying global importance-based pruning directly to autonomous driving scenarios presents limitations. Dynamic objects may only appear in certain cameras for limited time. As shown in Figure \ref{fig:3_stage2_pruning_comparison}, a car may initially appear in the left camera as it approaches, only to disappear from view as it moves past. This leads to incomplete reconstructions when their Gaussians are mistakenly pruned due to limited visibility across camera views. To mitigate this issue, we apply pruning only to \textbf{background Gaussians}. For a Gaussian $G_i$ belonging to a background node $n$, the importance metric is defined as:

\begin{equation}
I_i^{\text{bg}} = \frac{1}{M} \sum_{m=1}^M \sum_{j=1}^{K_m} \frac{w_{ij}^{(m)}}{S_i^{(m)}},
\end{equation}
where $M$ is the total number of training images, and $I_i^{\text{bg}}$ represents the normalized importance of $G_i$ across all images.

\subsection{Implementation Details}

We apply node-specific regularization coefficients and minimum thresholds to control pruning strength across different scene components. For static background nodes, we employ aggressive pruning with a regularization coefficient of $\lambda_{\text{bg}} = 2.50 \times 10^{-3}$ and maintain a minimum threshold of $K_{\text{min,bg}} = 40{,}000$ Gaussians. Dynamic rigid nodes (vehicles) use $\lambda_{\text{rigid}} = 1.00 \times 10^{-4}$ and $K_{\text{min,rigid}} = 10{,}000$, while deformable nodes adopt $\lambda_{\text{deform}} = 1.00 \times 10^{-5}$ and $K_{\text{min,deform}} = 5{,}000$ to preserve local deformation details. For SMPL nodes representing human actors, we employ $\lambda_{\text{SMPL}} = 1.00 \times 10^{-5}$ with a higher threshold of $K_{\text{min,SMPL}} = 50{,}000$ to maintain articulated motion fidelity.

%% file: sec/3_tables/05_experiment/1_comparison_waymo.tex
\begin{table*}[t]
\centering
\caption{Comparative performance of our framework and baseline approaches on the Waymo-NOTR dataset. M denotes million Gaussian primitives.}
\label{table1:performance_comparison}
\small
\setlength{\tabcolsep}{1pt}
\resizebox{1.0\textwidth}{!}{
\begin{tabular}{c c c c c c c c c c c c c c c}
\toprule
& \multicolumn{6}{c}{\textbf{Scene Reconstruction}} & \multicolumn{6}{c}{\textbf{Novel View Synthesis}} & \\
\cmidrule(lr){2-7} \cmidrule(lr){8-13}
\textbf{Methods} & \multicolumn{2}{c}{Full Image} & \multicolumn{2}{c}{Human} & \multicolumn{2}{c}{Vehicle} & \multicolumn{2}{c}{Full Image} & \multicolumn{2}{c}{Human} & \multicolumn{2}{c}{Vehicle} & {\textbf{\# Gauss↓}} & {\textbf{\# FPS↑}} \\
\cmidrule(lr){2-3} \cmidrule(lr){4-5} \cmidrule(lr){6-7} \cmidrule(lr){8-9} \cmidrule(lr){10-11} \cmidrule(lr){12-13}
& \textbf{PSNR↑} & \textbf{SSIM↑} & \textbf{PSNR↑} & \textbf{SSIM↑} & \textbf{PSNR↑} & \textbf{SSIM↑} & \textbf{PSNR↑} & \textbf{SSIM↑} & \textbf{PSNR↑} & \textbf{SSIM↑} & \textbf{PSNR↑} & \textbf{SSIM↑} & \\
\midrule
EmerNeRF~\cite{2023_11_3-EMERNERF-EMERGENT_SPATIAL_TEMPORAL_SCENE_DECOMPOSITION_VIA_SELF-SUPERVISION} & 31.93 & 0.902 & 22.88 & 0.578 & 24.65 & 0.723 & 29.67 & 0.883 & 20.32 & 0.454 & 22.07 & 0.609 & - & - \\
3DGS~\cite{2023_8_08-3dgs_for_real_time_radiance_field_rendering} & 26.00 & 0.912 & 16.88 & 0.414 & 16.18 & 0.425 & 25.57 & 0.906 & 16.62 & 0.387 & 16.00 & 0.407 & - & - \\
HUGS~\cite{2024_3_19-HUGS-Holistic_Urban_3D_Scene_Understanding_via_Gaussian_Splatting} & 28.26 & 0.923 & 16.23 & 0.404 & 24.31 & 0.794 & 27.65 & 0.914 & 15.99 & 0.378 & 23.27 & 0.748 & - & - \\
DeformGS~\cite{yang2023deformable3dgaussianshighfidelity} & 27.97 & 0.923 & 17.23 & 0.429 & 19.14 & 0.544 & 26.47 & 0.884 & 16.84 & 0.391 & 18.21 & 0.487 & 0.64M & 37.08 \\
PVG~\cite{2024_3_20-PVG-periodic_vibration_gaussian-dynamic_urban_scene_reconstruction} & 32.68 & 0.941 & 24.96 & 0.726 & 24.36 & 0.763 & 28.73 & 0.881 & 21.95 & 0.565 & 21.43 & 0.617 & 1.51M & 9.13 \\
StreetGS~\cite{2024_01_02-street_gaussian-modelling_dynamic_urban_scenes_with_gs} & 28.73 & 0.932 & 16.54 & 0.401 & 26.46 & 0.848 & 27.02 & 0.887 & 16.27 & 0.368 & 23.99 & 0.761 & 0.87M & 21.60 \\
OmniRe~\cite{2024_8_29-omnire} & \textbf{34.26} & \textbf{0.956} & \textbf{26.99} & \textbf{0.825} & \textbf{27.79} & \textbf{0.886} & \textbf{29.86} & \textbf{0.900} & \textbf{23.16} & \textbf{0.674} & \textbf{24.52} & \textbf{0.786} & 1.55M & 46.15 \\
\midrule
StreetGS+Ours & 28.42 & 0.924 & 16.51 & 0.392 & 26.33 & 0.847 & 27.04 & 0.889 & 16.18 & 0.362 & 23.72 & 0.753 & \textbf{0.29M} & 57.66 \\
OmniRe+Ours & 34.05 & 0.952 & 26.88 & 0.818 & 27.48 & 0.878 & 29.75 & 0.897 & 23.15 & 0.667 & 24.49 & 0.782 & 0.46M & \textbf{80.22}\\
\bottomrule
\end{tabular}
}
\end{table*}

%% file: sec/3_tables/05_experiment/2_comparison_nuscenes.tex
\begin{table*}[t]
\centering
\caption{Comparative performance of our framework and baseline approaches on the NuScenes dataset. M denotes million Gaussian primitives.}
\label{table:nuscenes_performance_comparison}
\small
\setlength{\tabcolsep}{1pt}
\resizebox{1.0\textwidth}{!}{
\begin{tabular}{c c c c c c c c c c c c c c c}
\toprule
& \multicolumn{6}{c}{\textbf{Scene Reconstruction}} & \multicolumn{6}{c}{\textbf{Novel View Synthesis}} & \\
\cmidrule(lr){2-7} \cmidrule(lr){8-13}
\textbf{Methods} & \multicolumn{2}{c}{Full Image} & \multicolumn{2}{c}{Human} & \multicolumn{2}{c}{Vehicle} & \multicolumn{2}{c}{Full Image} & \multicolumn{2}{c}{Human} & \multicolumn{2}{c}{Vehicle} & {\textbf{\# Gauss↓}} & {\textbf{\# FPS↑}} \\
\cmidrule(lr){2-3} \cmidrule(lr){4-5} \cmidrule(lr){6-7} \cmidrule(lr){8-9} \cmidrule(lr){10-11} \cmidrule(lr){12-13}
& \textbf{PSNR↑} & \textbf{SSIM↑} & \textbf{PSNR↑} & \textbf{SSIM↑} & \textbf{PSNR↑} & \textbf{SSIM↑} & \textbf{PSNR↑} & \textbf{SSIM↑} & \textbf{PSNR↑} & \textbf{SSIM↑} & \textbf{PSNR↑} & \textbf{SSIM↑} & \\
\midrule
DeformGS & \textbf{32.31} & 0.924 & 31.76 & 0.900 & 28.18 & 0.864 & \textbf{25.01} & \textbf{0.728} & \textbf{24.02} & \textbf{0.570} & 20.96 & \textbf{0.574} & 0.43M & 278.80 \\
StreetGS & 32.06 & 0.928 & 31.42 & 0.901 & \textbf{29.68} & 0.915 & 24.29 & 0.698 & 23.42 & 0.546 & \textbf{21.05} & 0.557 & 0.72M & 117.80 \\
OmniRe & 32.14 & \textbf{0.929} & \textbf{32.14} & \textbf{0.917} & 29.72 & \textbf{0.916} & 24.24 & 0.696 & 23.38 & 0.551 & 21.01 & 0.555 & 0.73M & 132.56 \\
\midrule
StreetGS+Ours & 31.08 & 0.913 & 30.08 & 0.868 & 28.56 & 0.897 & 24.06 & 0.698 & 23.21 & 0.544 & 20.90 & 0.558 & \textbf{0.26M} & \textbf{461.17} \\
OmniRe+Ours & 31.37 & 0.918 & 31.16 & 0.900 & 28.80 & 0.902 & 24.05 & 0.696 & 23.28 & 0.549 & 20.96 & 0.551 & 0.38M & 435.85 \\
\bottomrule
\end{tabular}
}
\end{table*}

%% file: sec/1_main/05_experiment.tex
\section{Experiments}
In this section, we comprehensively evaluate our SparseStreet framework on autonomous driving datasets.

\subsection{Datasets and Metrics} 
\label{subsec:Datasets and Metrics}

\noindent\textbf{Dataset and Baselines.} To thoroughly assess the effectiveness of our proposed compression framework, we conduct experiments on two large-scale autonomous driving datasets. For the Waymo Open Dataset~\cite{sun2020_waymo}, following the experimental setup in OmniRe~\cite{2024_8_29-omnire}, we select 8 highly dynamic scenes characterized by diverse foreground objects, including pedestrians and cyclists, as well as complex static backgrounds. Similarly, for the nuScenes dataset~\cite{caesar2020nuscenes}, following the experimental protocol of NeuRAD~\cite{2023_11_neurad}, we evaluate on 8 representative scenes (scene IDs: 39, 54, 61, 66, 104, 108, 122, 176) that encompass diverse urban driving scenarios with varying levels of dynamic content.

\noindent\textbf{Hardware and Training Configuration.} All experiments were conducted using a single NVIDIA A800 GPU. Our two-stage training process begins with node-aware pruning from the start until step 24,000, applying different regularization strengths using the masking loss $\mathcal{L}_m$. At step 28,000, we apply background pruning, removing all background Gaussian primitives with zero importance scores while retaining all others.

\noindent\textbf{Evaluation Metrics.} To evaluate reconstruction quality and novel view synthesis (NVS) performance, we report PSNR and SSIM as metrics for both overall scene quality and vehicle-specific fidelity. Additionally, LPIPS is computed as a perceptual quality metric. Novel view synthesis is evaluated on every 10th frame. In addition, for novel trajectory synthesis evaluation, we employ the FID metric to assess the perceptual quality of generated views along unseen camera trajectories. We further report the number of Gaussian primitives (\# Gauss) to assess compression efficiency and frames-per-second (FPS) to evaluate rendering speed.

\begin{figure*}[t!]
\centering
\includegraphics[width=\textwidth]{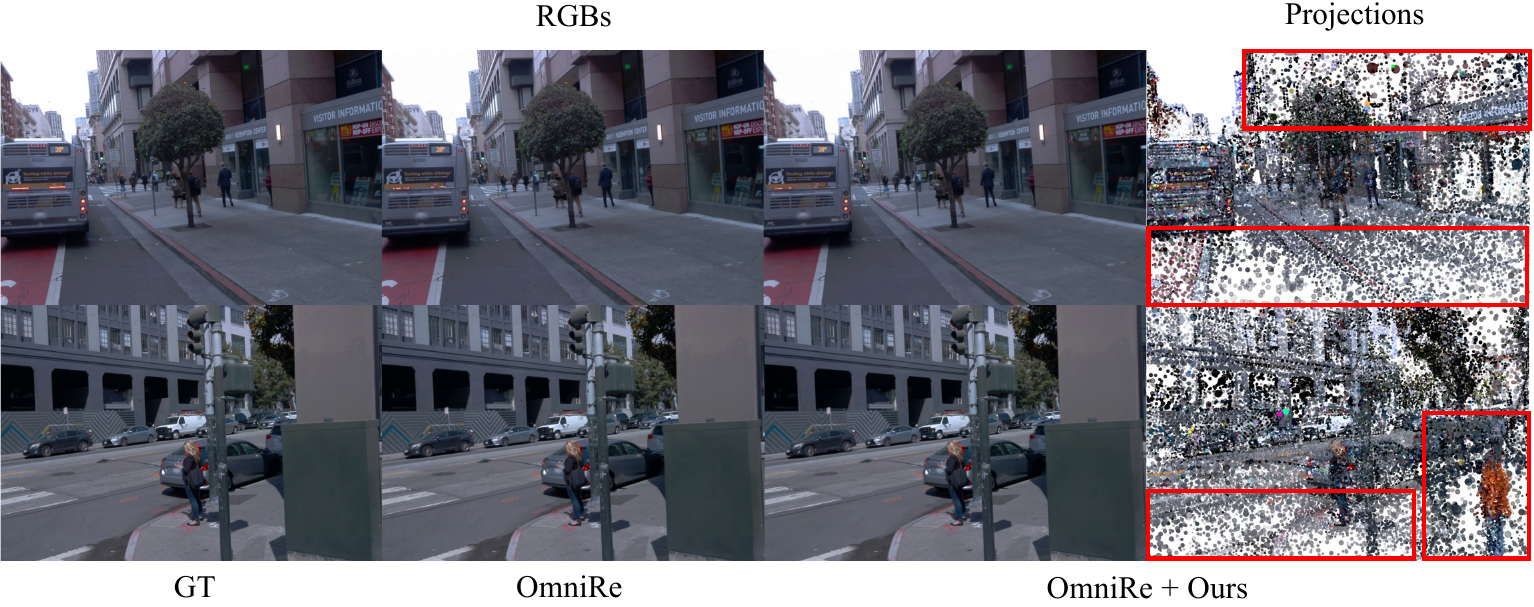}
\captionof{figure}{Qualitative comparisons of ground truth (GT), OmniRe, and OmniRe + Ours. The fourth column shows Gaussian projections of our method. Red boxes highlight that static elements (ground plane, buildings) can be effectively represented with fewer Gaussians, while dynamic objects (person) maintain dense representation for better quality.}
\label{fig:2_ground_comparisons}
\end{figure*}

\subsection{Comparative Results}
\label{subsec:Comparative Results}

To validate the effectiveness of our proposed SparseStreet framework, we conduct experiments on two state-of-the-art baselines: \textbf{OmniRe} \cite{2024_8_29-omnire} and \textbf{StreetGS} \cite{2024_01_02-street_gaussian-modelling_dynamic_urban_scenes_with_gs}. Since prior works did not report the number of Gaussian primitives (\# Gauss) and frames-per-second (FPS), we reimplemented DeformGS, PVG, StreetGS, and OmniRe methods to obtain these results using the official DriveStudio repository\footnote{\url{https://github.com/ziyc/drivestudio}}.

\input{sec/3_tables/05_experiment/4_global_pruning_vs_background_pruning}

\begin{figure*}[t]
\centering
\includegraphics[width=\textwidth]{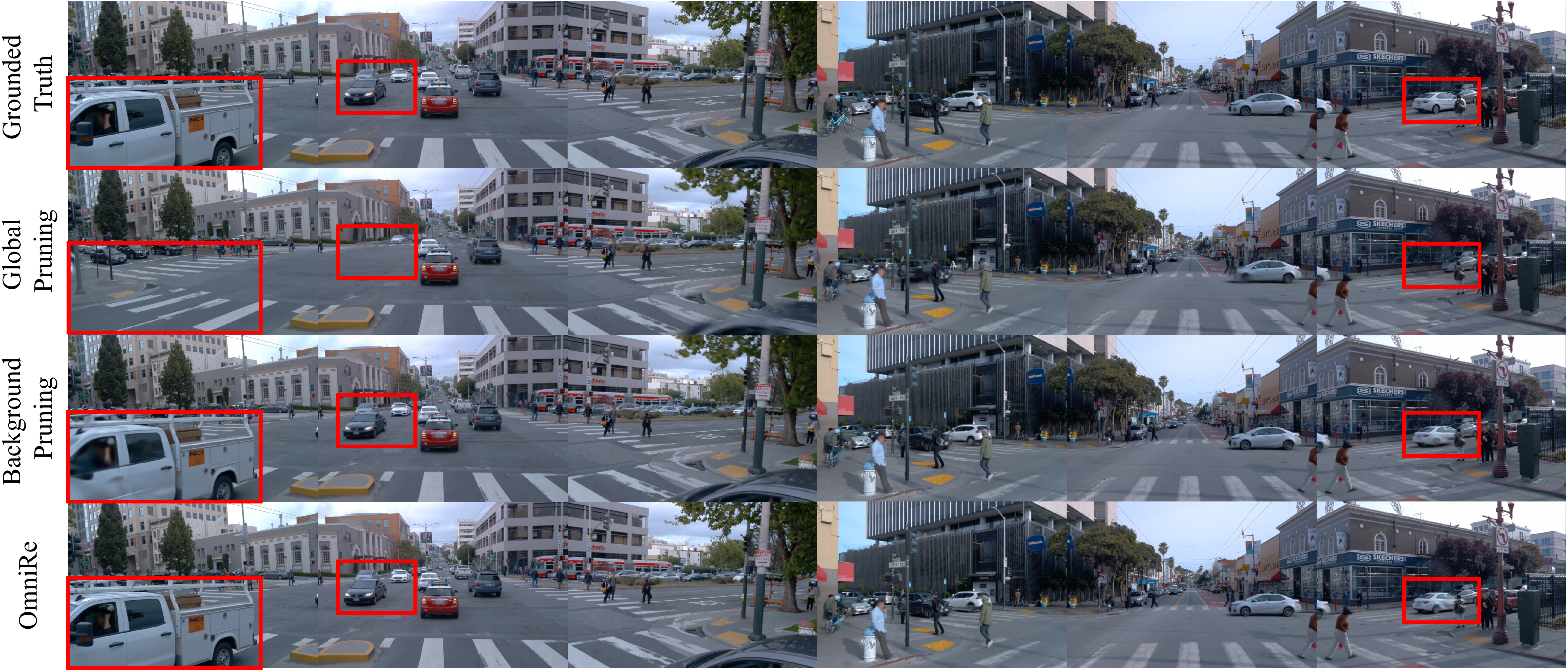}
\caption{Comparison of different pruning strategies on dynamic objects across three camera views. First row: Ground truth images. Second row: Results using global pruning, showing missing parts of moving vehicles due to their limited temporal presence in specific views. Third row: Our background pruning approach, which preserves the integrity of dynamic objects while effectively compressing static scene elements. Fourth row: OmniRe results.}
\label{fig:3_stage2_pruning_comparison}
\end{figure*}

Tab.~\ref{table1:performance_comparison} compares the performance of SparseStreet with several baselines on the Waymo dataset~\cite{sun2020_waymo}. Our method achieves a significant reduction in Gaussian primitives (\# Gauss) and a substantial increase in FPS, making it more efficient for real-time rendering in autonomous driving applications. When integrated with OmniRe, our approach reduces the number of Gaussian primitives by approximately 3× with only minimal quality degradation. Similarly, when applied to StreetGS, our method significantly decreases the Gaussian count while simultaneously increasing the rendering frame rate. While there is a minimal impact on visual quality metrics, our approach maintains competitive novel view synthesis (NVS) performance in certain metrics. These results highlight SparseStreet's ability to effectively balance compression efficiency, rendering speed, and visual quality. Most notably, our approach enables real-time rendering performance, achieving up to 80.22 FPS when combined with OmniRe. Fig.~\ref{fig:teaser} shows Gaussian projections before and after pruning across three camera views. Our approach effectively removes redundant primitives in background regions while preserving details in dynamic objects, demonstrating effective prioritization under aggressive pruning conditions.

Tab.~\ref{table:nuscenes_performance_comparison} presents results on the NuScenes dataset. Our framework demonstrates consistent performance improvements across both datasets. When applied to StreetGS on NuScenes, our method achieves a remarkable 2.8× reduction in Gaussian primitives while increasing FPS by 3.9×. Similarly, OmniRe+Ours maintains competitive reconstruction quality while reducing Gaussian count by approximately 1.9× and achieving 3.3× faster rendering compared to the baseline.

\subsection{Novel Trajectory Synthesis}
\label{subsec:Novel Trajectory Synthesis}

We also evaluate our framework's ability to render novel trajectories for autonomous driving applications on the Waymo dataset. As shown in Tab.~\ref{tab:lane_change}, our method achieves comparable FID scores to the baseline across different lateral offsets while using significantly fewer Gaussian primitives, demonstrating preserved generalization capability.

\input{sec/3_tables/05_experiment/3_novel_trajectory_synthesis}

\subsection{Ablation Studies}
\label{subsec:Ablation Studies}

We conduct comprehensive ablation studies to assess the contribution of individual components in our framework, as shown in Tab.~\ref{table2:ablation}. We evaluate the impact of applying \textbf{Node-aware Pruning} and \textbf{Background Pruning} individually and in combination.

\subsubsection{Component Analysis}
As shown in Tab.~\ref{table2:ablation}, node-aware pruning alone achieves high reconstruction quality but maintains a relatively large number of Gaussians. Background pruning significantly reduces the Gaussian count while preserving competitive quality. The combination of both approaches achieves the optimal balance between compression efficiency and reconstruction fidelity.

\input{sec/3_tables/05_experiment/5_ablation_study}

\subsubsection{Background Pruning Analysis} 

We compare our background-only pruning strategy with global importance-based pruning applied to all Gaussians. As shown in Tab.~\ref{tab3:background_redundancy_removal}, our approach achieves superior reconstruction quality, with PSNR improving by +1.13 compared to global pruning while maintaining comparable compression efficiency. Overall, our comprehensive evaluations confirm that SparseStreet's component-wise design effectively balances compression efficiency and visual quality across diverse autonomous driving scenarios.

%% file: sec/3_tables/05_experiment/4_global_pruning_vs_background_pruning.tex
\begin{table}[H]
\centering
\caption{Comparison of different pruning strategies on the Waymo Open dataset. M denotes million Gaussian primitives.}
\label{tab3:background_redundancy_removal}
\small
\resizebox{\columnwidth}{!}{%
\begin{tabular}{@{}c@{\hspace{6pt}}c@{\hspace{6pt}}c@{\hspace{6pt}}c@{\hspace{6pt}}c@{\hspace{6pt}}c@{\hspace{6pt}}c@{}}
\toprule
\multirow{2}{*}[-\dimexpr\ht\strutbox/2]{\textbf{Method}} & \multicolumn{3}{c}{\textbf{Full Image}} & \textbf{Human} & \textbf{Vehicle} & \multirow{2}{*}[-\dimexpr\ht\strutbox/2]{\textbf{\# Gauss}} \\
\cmidrule(l{0pt}r{7pt}){2-4} \cmidrule(l{-1pt}r{6pt}){5-5} \cmidrule(l{-1pt}r{6pt}){6-6}
 & \textbf{PSNR} & \textbf{SSIM} & \textbf{LPIPS} & \textbf{PSNR} & \textbf{PSNR} & \\
\midrule
Global Pruning & 32.18 & 0.940 & 0.077 & 26.07 & 23.89 & 0.33M \\
Background Pruning & 33.32 & 0.945 & 0.073 & 26.54 & 26.66 & 0.39M \\
\bottomrule
\end{tabular}%
}
\end{table}

%% file: sec/3_tables/05_experiment/3_novel_trajectory_synthesis.tex
\begin{table}[h]
\centering
\caption{Novel trajectory synthesis on Waymo dataset, showing FID scores for different lateral offsets from the original trajectory.}
\label{tab:lane_change}
\small
\begin{tabular}{@{}c@{\hspace{15pt}}c@{\hspace{15pt}}c@{\hspace{15pt}}c@{}}
\toprule
\multirow{2}{*}[-\dimexpr\ht\strutbox/2]{\textbf{Method}} & \multicolumn{3}{c}{\textbf{FID↓ (Avg. of Left/Right)}} \\
\cmidrule(l{0pt}r{0pt}){2-4} 
 & \textbf{1m} & \textbf{2m} & \textbf{3m} \\
\midrule
OmniRe & 43.62 & 61.56 & 75.14 \\
OmniRe + Ours & \textbf{46.54} & \textbf{66.28} & \textbf{80.36} \\
\bottomrule
\end{tabular}
\label{tab:novel_traj}
\end{table}

%% file: sec/3_tables/05_experiment/5_ablation_study.tex
\begin{table}[H]
\centering
\caption{Ablation study on the Waymo Open dataset showing the impact of different components in our framework. M denotes million Gaussian primitives.}
\vspace{-1mm}  
\label{table2:ablation}
\small
\resizebox{\columnwidth}{!}{%
\begin{tabular}{ccccccc}
\toprule
\multirow{2}{*}[-\dimexpr\ht\strutbox/2]{\textbf{Method}} & \multicolumn{3}{c}{\textbf{Full Image}} & \textbf{Human} & \textbf{Vehicle} & \multirow{2}{*}[-\dimexpr\ht\strutbox/2]{\textbf{\# Gauss}} \\
\cmidrule(lr){2-4} \cmidrule(lr){5-5} \cmidrule(lr){6-6}
& \textbf{PSNR} & \textbf{SSIM} & \textbf{LPIPS} & \textbf{PSNR} & \textbf{PSNR} & \\
\midrule
Full Model & 33.35 & 0.945 & 0.072 & 26.50 & 26.88 & 0.32M \\
w/o Node Pruning & 33.32 & 0.945 & 0.073 & 26.54 & 26.66 & 0.39M \\
w/o Background Pruning & 33.71 & 0.951 & 0.063 & 26.61 & 27.03 & 0.65M \\
\bottomrule
\end{tabular}%
}
\vspace{-2mm}
\end{table}

%% file: sec/1_main/06_conclusion.tex
\section{Conclusion}
In this paper, we present SparseStreet, a compression framework dedicated to street Gaussian splatting representations. By introducing node-aware pruning and background compression, our approach addresses the storage challenges in dynamic street scene reconstruction. Comprehensive experiments on the Waymo and nuScenes datasets demonstrate that SparseStreet significantly reduces Gaussian primitives while maintaining high visual fidelity, enabling real-time rendering performance for practical autonomous driving applications.

While SparseStreet is currently developed under a supervised setting using scene graphs with semantic annotations, recent advances in self-supervised dynamic scene modeling have exhibited promising capabilities in decomposing static and dynamic components. We believe that integrating SparseStreet with self-supervised street Gaussian frameworks is a promising direction for future work, enabling efficient compression without relying on costly annotations.

%% file: sec/1_main/10_acknowledgments.tex

\begin{acks}
This work was supported by the National Natural Science Foundation of China (NSFC) under Grant No. W2542034.
\end{acks}

%% file: sec/1_main/11_reference.tex


\bibliographystyle{ACM-Reference-Format}
\bibliography{main}